\documentclass{article}

\usepackage[preprint]{neurips_2026}

\usepackage[utf8]{inputenc} %
\usepackage[T1]{fontenc}    %
\usepackage{hyperref}       %
\usepackage{url}            %
\usepackage{booktabs}       %
\usepackage{amsfonts}       %
\usepackage{nicefrac}       %
\usepackage{microtype}      %
\usepackage{xcolor}         %
\usepackage{comment}
\usepackage{placeins}

\usepackage{csquotes}
\usepackage{listings}       %
\usepackage{subcaption}
\usepackage{graphicx}
\newcommand{\code}[1]{%
  \colorbox{gray!15}{\texttt{#1}}%
}

\title{A Shared Subcircuit Lets LLMs Count Down Across Tasks}

\author{%
  Jacob Dunefsky \\
  Yale University\\
  \texttt{jacob.dunefsky@yale.edu} \\
  \And
  Wes Gurnee \\
  Anthropic \\
  \And
  Emmanuel Ameisen \\
  Anthropic \\
}

\begin{document}

\maketitle

\begin{abstract}
Writing a sentence of exactly twelve words; ending a DNA sequence at the right codon; formatting an ASCII table.
These are all tasks that language models can do that requires tracking how many tokens remain before a target. In this work, we identify in Llama-3.1-70B-Instruct a general mechanism for performing these tasks: a \enquote{countdown subcircuit} that compares the current position to a goal length and estimates the time remaining until then. We first isolate a countdown subcircuit in a controlled setting, in which the model is tasked with writing a fixed-length sentence ending in a specified word. 
We then investigate the geometry of the representations used by the subcircuit, and find that the subcircuit uses an identical motif previously identified in a frontier LLM on a separate task, thus suggesting that this motif is shared across models.
Finally, we use unsupervised probing on a natural language dataset to find a variety of other tasks where this subcircuit is used, including tasks where the goal length is inferred from context rather than explicitly stated.
Our work suggests that reverse-engineering subcircuits allows us to understand how behaviors generalize from a single example to many different tasks and even models.
\end{abstract}

\section{Introduction}

Mechanistic interpretability seeks to reverse-engineer complex neural networks, such as large language models (LLMs), in order to allow humans to reason about the algorithms which they implement.
Just as reverse-engineering a complex computer program entails understanding individual subroutines, so too does reverse-engineering neural networks require understanding \textit{subcircuits}: subgraphs of the neural network's computational graph that are responsible for specific behaviors. Initial pioneering work in LLM subcircuit interpretability, such as that of \citet{wang2022interpretability}, focused on taking a specific well-defined task on a narrow distribution of inputs and finding a subcircuit that implements the required computation for that task on those inputs.

And as the field has matured, there has been more research that has investigated subcircuits that are \textit{reused} across different tasks. For example, \citet{merullo2024circuit} found that the subcircuit identified by \citet{wang2022interpretability} is also responsible for computing a seemingly different task which had been previously considered in the literature~\citep{srivastava2023beyond}. More recently, \citet{feucht2026arithmeticwildllamauses} found a \enquote{base-10 addition} subcircuit that is used for performing arithmetic across various cyclic settings including adding raw numbers, working with months, weekdays, and hours. And in a more naturalistic setting, \citet{gurnee2025when} identified a subcircuit responsible for inserting linebreaks at the correct positions in line-wrapped text in varied contexts.

In our work, we deeply investigated a specific example of such a reused subcircuit: a \enquote{countdown} subcircuit that Llama-3.1-70B-Instruct uses to track the amount of time remaining until a target goal length is reached. Notably, we took a \textit{bottom-up approach} to understanding the full generality of contexts in which this subcircuit is used: after first reverse-engineering this subcircuit on a single task, we used unsupervised probing methods to find a wide variety of other contexts \textit{in an unlabeled corpus} that make use of this subcircuit. 

Note that concurrent work by \citet{merzouk2026much} also investigates probing LLMs' internal states to predict the length of their responses on chat datasets. Compared to theirs, our work focuses on using causal interventions to reverse-engineer the subcircuit and geometry used by the model to process length information; the use of probes for the \textit{unsupervised discovery} of other settings and examples where this subcircuit is used is also a unique focus of our work.

\begin{figure}
    \centering
    \includegraphics[width=0.85\textwidth]{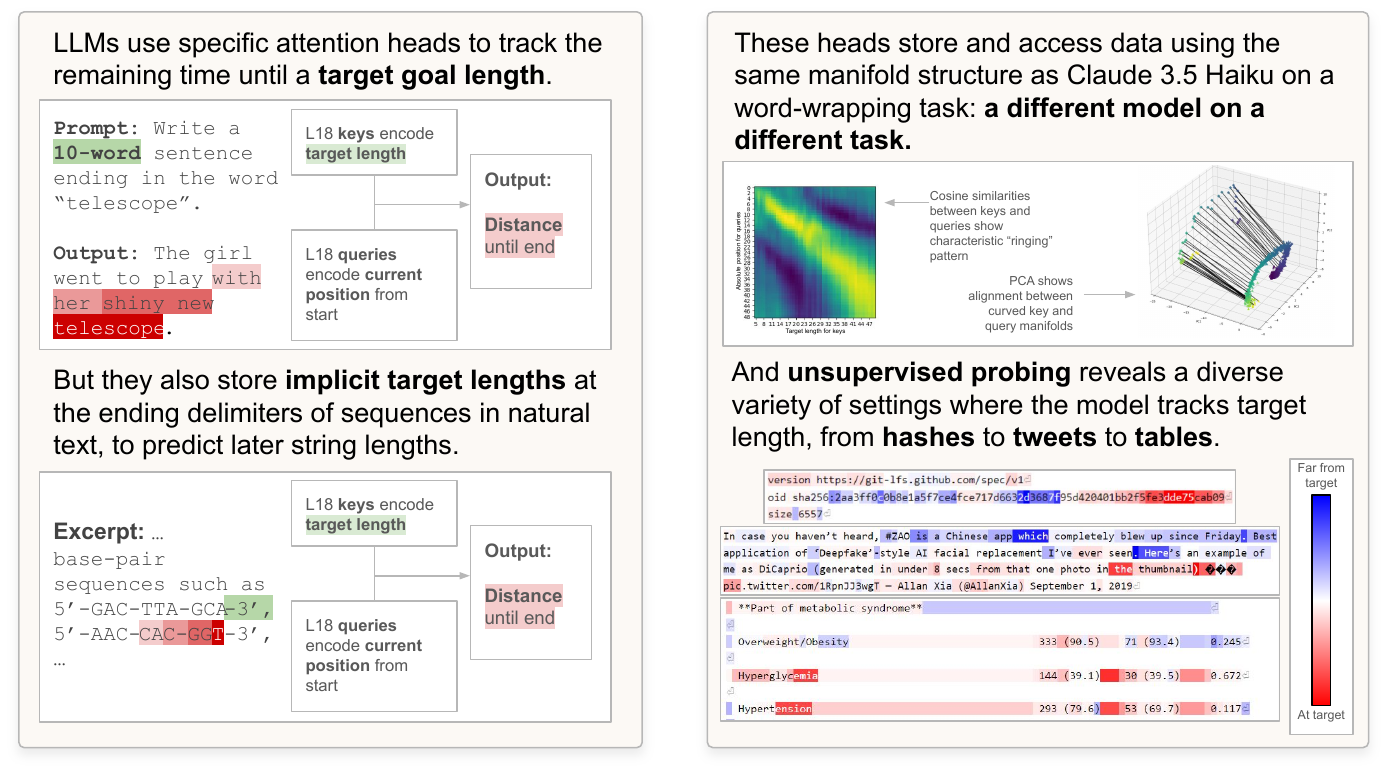}
    \caption{We find a small subcircuit in an LLM used for tracking target goal length, and further find that this subcircuit is used across tasks and models.}
\end{figure}

\FloatBarrier

Our contributions are the following:
\begin{itemize}
    \item (Sec.~\ref{sec:identifyingCircuit}) We reverse-engineer a specific sentence planning task, where the model is instructed to output a sentence of a given length, in order to find \textbf{countdown attention heads} responsible for keeping track of how much longer the model has in the sentence.
    \item (Sec.~\ref{sec:mechanism}) We further analyze how the countdown subcircuit is used \textit{in cases where a target length is not explicitly given in the prompt}, with the model inferring the target length. We find that the model stores implicit length information at ending delimiters of strings.
    Furthermore, we find that the geometry of this length information, along with later positional information in prompts, is largely the same as that found previously in Claude 3.5 Haiku on a line-wrapping task, \textbf{suggesting that this countdown subcircuit motif is present across models}.
    \item (Sec.~\ref{sec:taskUniversality}) We perform unsupervised probing with the output of the countdown heads, and find that this subcircuit is used in many different tasks, from predicting the end of tweets embedded in news articles to formatting ASCII tables.
\end{itemize}

\section{Identifying \enquote{countdown} heads with a sentence planning task}
\label{sec:identifyingCircuit}

\subsection{Sentence planning task}
\label{sec:sentencePlanning}
In order to understand how models plan ahead in their outputs, we chose to study a specific task
which we call the \enquote{sentence planning task}. In this task, the model is instructed to write a sentence of a given length ending in a given word. For example, one prompt is

\begin{lstlisting}
    Write a 10-word sentence ending in the word "telescope".
\end{lstlisting}

In general, each prompt contains a \textbf{target length token} (the target length of the sentence; in the example prompt, \code{10}) and a \textbf{target word token} (in the example prompt, \code{telescope}). Additionally, after the target length token is always the token \code{-word}, which we refer to as the \textbf{length unit token}.

The model can perform this task well: for target lengths less than 11 words, the model's outputs are (almost) all exactly the same length as the target, with errors slowly increasing for target lengths higher than 11 words; at a target length of 24 words, the median error is less than 4 words. Given that the model performs well, the natural next question to ask is how it does so. This is what we investigate in the following sections.

\subsection{Causal countdown representations are localized at specific layers}

We seek to localize the representations that the model uses for determining when to end the output sentence. To do this, we begin by performing residual stream patching as follows. We consider pairs of prompts with the same target word, but differing in the target lengths: the prompt that we are patching into (the \enquote{destination prompt}) has target length $n_{dst}$, while the prompt that we are patching from (the \enquote{source prompt}) has target length $n_{src}$. We filter for prompt pairs such that $n_{dst} > n_{src}$, and such that the first $n_{src}-1$ words are the same in the model's output on both the source prompt and the destination prompt.

\begin{figure}
    \centering
    \includegraphics[width=0.5\linewidth]{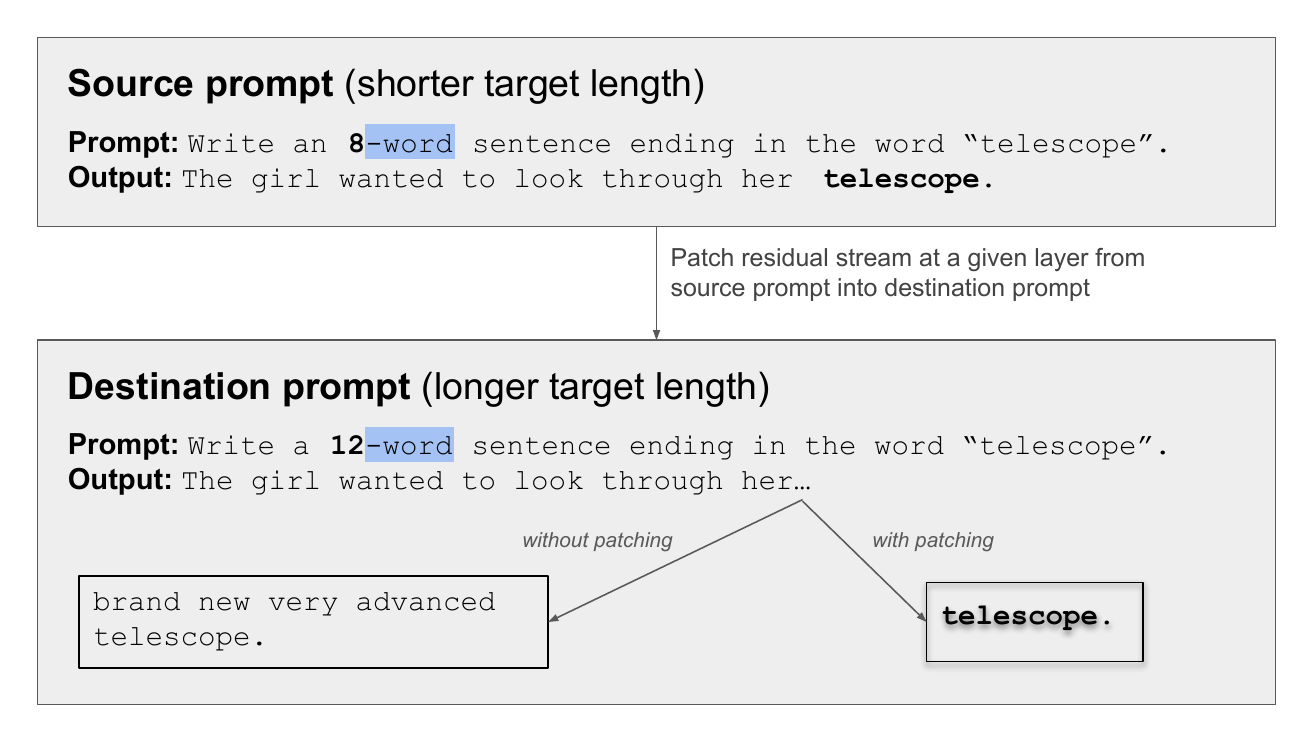}
    \caption{A high-level description of the patching setup for localizing length information.}
    \label{fig:layerLocalizationSetup}
\end{figure}

We truncate the model's output on the destination prompt before the word at position $n_{src}$. Then, at each layer, we patch the activations from the patching prompt into the residual stream at the target length token and (separately) at the length unit token. We record the logit of the target word after patching at each layer. If this logit is high, then this means that the patched model outputs a sentence of length $n_{src}$ rather than $n_{dst}$, implying that the patched activations at that layer are causally sufficient for changing the length of the sentence. Figure~\ref{fig:layerLocalizationSetup} provides a high-level schematic of the patching experiment.

\begin{figure*}[t]
    \centering
    \begin{minipage}{0.48\textwidth}
        \centering
        \includegraphics[width=\linewidth]{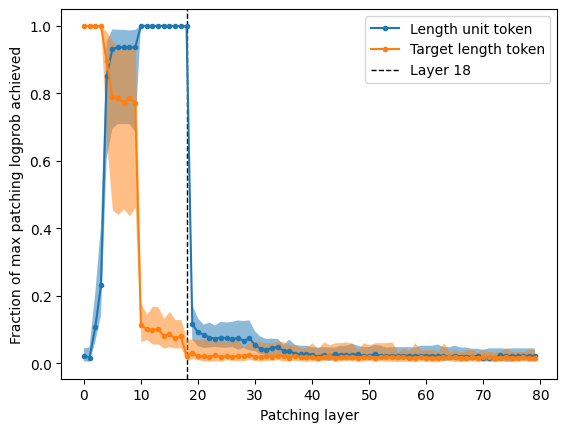}
        \captionof{figure}{Results of patching at different residual stream layers. Solid lines denote medians, shaded areas denote values between the first and third quartile. Y-axis corresponds to the fraction of the max logit of the target word achieved when patching at any layer (i.e. 0 corresponds to the minimum value for each prompt and 1 corresponds to the maximum).}
        \label{fig:lengthUnitResidPatching}
    \end{minipage}
    \hfill
    \begin{minipage}{0.48\textwidth}
        \centering
        \includegraphics[width=\linewidth]{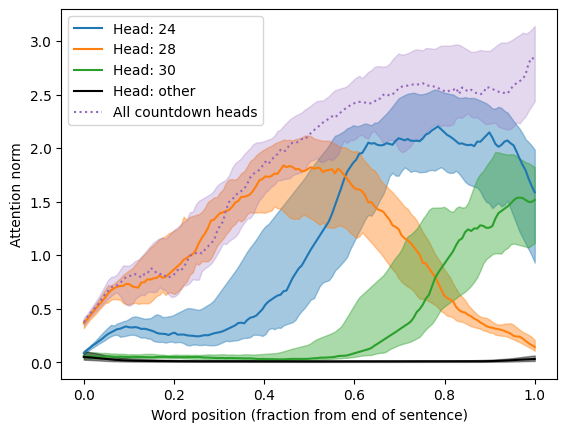}
        \captionof{figure}{Attention norms for layer 18 attention outputs from the length unit token to downstream tokens. Solid lines denote medians; the shaded area denotes the range from the first quartile to the third quartile. The x-axis normalizes token positions (for outputs of different lengths) by mapping them to the fraction of the sentence at which they are located.}
        \label{fig:attnNorms}
    \end{minipage}
\end{figure*}

\paragraph{Length info is last used at layer 18 at the length unit token.} Results are given in Figure~\ref{fig:lengthUnitResidPatching}. We see that when we patch into the length unit token starting at around layer 4, the target word's logit increases drastically, indicating that length information is moved to this token starting at around layer 4. But after layer 18, there is a drastic drop in the patched target word's logit (to almost the same point as the unpatched logit). This suggests that the length information is last causally used (at this token) at layer 18: presumably, at this point, the relevant information is read from the length unit token into a future token, and from layer 19 onward, any information still contained in the length unit token is noncausal.

\subsection{Specific attention heads count down to the end of the sentence}

Because the previous results tell us that length information is no longer used at the length unit token past layer 18, this implies that there are attention heads at layer 18 that refer back to this information in order to determine when to end the output sentence. To identify these attention heads, we looked at the norms of the outputs of all layer 18 attention heads from the length unit token to downstream tokens. (That is to say, for each attention head, at each downstream token, we look at the norm of the contribution to the output of that attention head at that downstream token from the length unit token as the source token.)

Figure~\ref{fig:attnNorms} shows these attention norms at different positions in the output sentences, where for each output sentence, the position in the sentence is divided by the length of the sentence (in order to normalize sentences of different lengths).

\paragraph{There are three relevant heads, and they count distance until the target length.} The only heads with output norms noticeably above zero are L18H24, L18H28, and L18H30 (where the notation \enquote{L$x$H$y$} denotes attention head $y$ at layer $x$). The heads display a pattern of overlapping receptive fields: L18H28's output is larger relatively early in the sequence before fading away, L18H24's output starts getting large in the middle of the sequence, and L18H30's output is largest right at the end of the sequence. The combined effect of this for the total attention norm to be highest right before the target length.

Notably, these heads' total output norm is highest right before the target length, and this pattern is consistent across sentences of different target lengths---implying that the output of these heads is indeed \enquote{counting down} until the target length. Furthermore, the total norm smoothly increases as the target length is approached, rather than suddenly spiking at the target length, suggesting that these heads are continuously counting down. We thus consider it justified to refer to these three heads as \textbf{countdown heads}.

\section{Understanding how the countdown subcircuit uses implicit length information}
\label{sec:mechanism}

Thus far, we have identified a \enquote{countdown subcircuit} used in the sentence planning setting, where three attention heads read information about a target length and keep track of the distance from the current position to the target length. However, in that setting, the target length is explicitly given in the prompt. But models can keep track of length information even when not given explicitly. This raises a new question: is this countdown subcircuit ever used in any cases where the target length is not explicitly given in the prompt?

We find that the answer is yes: \textit{there are cases where the model infers an implicit target length from context, which is then used in the countdown subcircuit}. In particular, we study a setting where prompts contain repeated strings of the same length, and find that the model treats the length of these strings as an implicit target length. In this setting, we have obtained the following results, which we will elaborate upon further:
\begin{enumerate}
    \item The model implicitly stores target length information at the ending delimiters of strings. This implicit length information is represented in a similar way as the explicit length information used in the sentence planning task.
    \item The countdown heads implement a motif previously identified in the linebreak mechanism in Claude 3.5 Haiku~\citep{gurnee2025when}, \textit{indicating cross-model universality of this motif}.
\end{enumerate}

\subsection{The repeated string setting}

We refer to the setting that we consider here as the \enquote{repeated string setting}. Broadly speaking, in this setting, the model processes prompts that contain multiple delimited strings of an identical length. There is freedom in what the type of string is, and what the delimiters are. For instance, one type of prompt might contain backtick-delimited base64 strings; another type of prompt might contain DNA base-pair sequences. Examples of such prompts are given in Figure~\ref{fig:repeatedStrings}.

\definecolor{lightblue}{RGB}{201,218,248}
\definecolor{lightred}{RGB}{234,153,153}

\begin{figure}
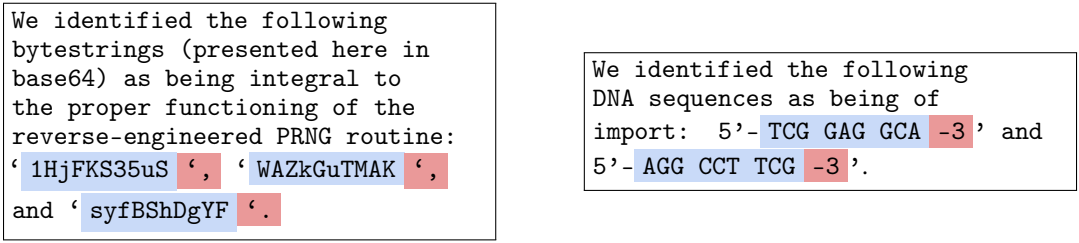

    \centering
    \begin{subfigure}{0.45\linewidth}
        \noindent\fbox{%
        \begin{minipage}{\textwidth}
            \texttt{We identified the following bytestrings (presented here in base64) as being integral to the proper functioning of the reverse-engineered PRNG routine: `\colorbox{lightblue}{1HjFKS35uS}\colorbox{lightred}{`,} `\colorbox{lightblue}{WAZkGuTMAK}\colorbox{lightred}{`,} and `\colorbox{lightblue}{syfBShDgYF}\colorbox{lightred}{`.}}
        \end{minipage}}%
    \end{subfigure}
    \hfill
    \begin{subfigure}{0.45\linewidth}
        \noindent\fbox{%
        \begin{minipage}{\textwidth}
            \texttt{We identified the following DNA sequences as being of import: 5'-\colorbox{lightblue}{TCG GAG GCA}\colorbox{lightred}{-3}' and 5'-\colorbox{lightblue}{AGG CCT TCG}\colorbox{lightred}{-3}'}.
        \end{minipage}}%
    \end{subfigure}
    \caption{Repeated-string prompt templates used in implicit length experiments. Left: an example base64 string prompt. Right: an example DNA sequence prompt. In both cases, the strings themselves are colored in light blue, and the ending delimiter \textit{tokens} are colored in light red.}
    \label{fig:repeatedStrings}
\end{figure}

As we will shortly see, in the repeated string setting, the model uses the length of previous strings in the prompt to predict when future strings should end---and the model stores this length information in the ending delimiters of strings.

\subsection{Causal investigation: target length information in string ending delimiters is causally important}
\label{sec:implicitCausal}

First, we aim to determine which tokens store implicit target length information that is \textit{causally important} for determining the model's prediction of the current string's length. One sensible ansatz is that information about the length of a string is stored in the token delimiting the end of that string. We perform a patching experiment in order to see if this ansatz is true. 

\paragraph{Setup} We make a prompt containing two randomly-generated length-$l_{dst}$ base64 strings delimited by backticks; we call this prompt the destination prompt. We also make a prompt containing two randomly-generated length-$l_{src}$ strings, called the source prompt. (In our experiments, $l_{dst}$ takes the values $\{30, 40, 50\}$, while $l_{src}$ ranges over the values from 10 to 24 (inclusive). Notice that all of the values of $l_{src}$ are less than all of the values of $l_{dst}$.) We then append to the destination prompt a third randomly-generated string of length $l_{dst}$; then, we patch the model and run it on the destination prompt, replacing all activations at the ending delimiter of the first two strings with the source prompt activations at the corresponding ending delimiters. We look at the next-token distribution of the patched model and record the first character position at which the most likely next token is the ending delimiter token---the predicted string length.

If all target length information is really stored in the ending delimiters of the string, then the predicted string length should be roughly equal to $l_{src}$.
Thus, we store the difference between $l_{src}$ and the predicted string length according to the patched model; lower is better. As a baseline, we run the \textit{unpatched} model on the \textit{destination} prompt; we expect the difference metric to be very high (because the previous strings in the destination prompt have length $l_{dst}$, not $l_{src}$). As a roofline, we run the \textit{unpatched} model on the \textit{source} prompt without patching; because the previous strings in the source prompt have length $l_{src}$, we expect the difference between $l_{src}$ and the predicted length to be very small.

For each $(l_{dst}, l_{src})$ pair, we perform five iterations of this experimental procedure.

\paragraph{Results} 

\begin{figure}
    \centering
    \includegraphics[width=0.95\linewidth]{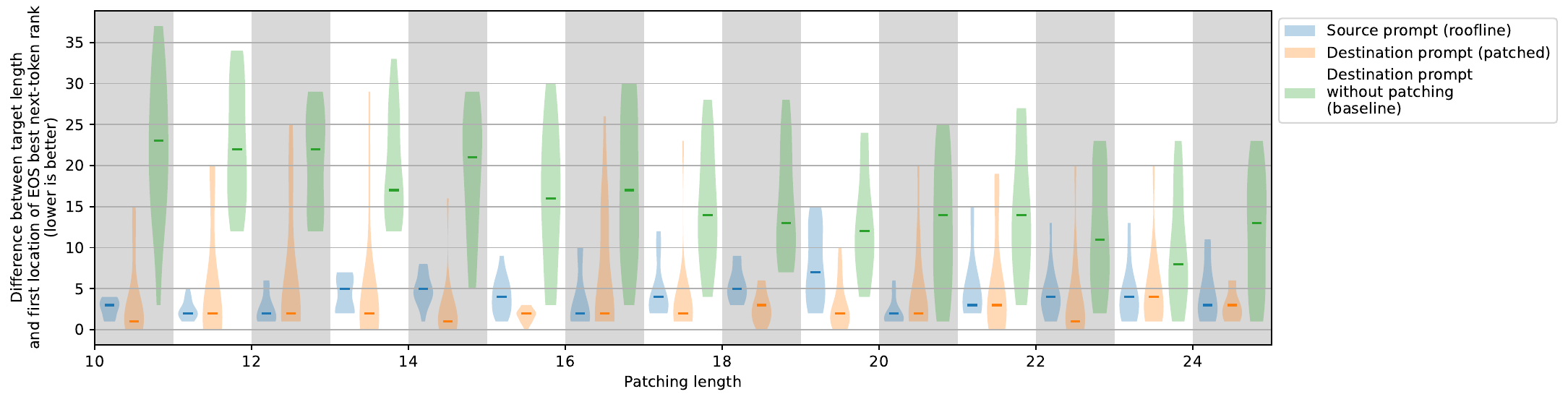}
    \caption{Difference between patched string lengths and target string lengths; lower is better. Solid lines denote medians.}
    \label{fig:delimiterAblation}
\end{figure}
Figure~\ref{fig:delimiterAblation} shows the results of this experiment. It is first worth noting that the roofline has low values for the difference metric, confirming that the model does indeed predict that the third string will have (approximately) the same length as the first two strings. But more interestingly, we see that patching the delimiter tokens on the destination prompt causes the model to predict that the third string has length $l_{src}$ (corresponding to the prompt that we are patching with) rather than $l_{dst}$ (corresponding to the original prompt), nearly to the same extent that the unpatched model run on the source prompt predicts that the third string has length $l_{src}$, and (usually) far below the baseline. This suggests that the activations of the ending delimiter tokens are causally sufficient for determining the length predicted by the model, strongly suggesting that implicit target length information is indeed stored in string ending delimiters.

\subsection{Observational investigation: implicit and explicit target lengths are represented similarly}

Having seen that implicit length information is indeed stored in the ending delimiters of target strings, we now want to better understand how this information is stored. In particular, is this information stored similarly across different tasks? And is implicit length information stored similarly to explicit length information?

To investigate this, we take an observational approach, looking at activations from prompts containing implicit length information along with prompts containing explicit length information. More specifically, we look at the key activations for the key-value head corresponding to all of the countdown heads\footnote{The model that we are investigating, Llama-3.1-70B-Instruct, uses grouped-query attention, such that there is one key-value head for every eight query heads. All of the countdown heads identified use the same key-value head.}, in order to test how well these activations can be used as length probes across the different prompt types.

We take these activations at the ending delimiters of strings in two separate \enquote{implicit length} prompt types: prompts containing base64 strings delimited by backticks, and prompts containing DNA sequences delimited by \code{5'-} and \code{-3'} delimiters. We also look at the activations at the token \code{-word} in prompts of the form \enquote{Write a $k$-word sentence} for varying values of $k$; this corresponds to explicit length information. We do this for prompts with varying target lengths.

Once we have obtained these activations, we take the pairwise cosine similarities of activations of different lengths, across different prompt types. If two prompt types store target length information relatively similarly, then we should expect to see the following relationship in the cosine similarities. Specifically, let Prompt A be a prompt with Prompt Type A and target length $l_A$. We compare the key activations from Prompt A to the key activations from prompts with Prompt Type B of different lengths. We should then see the following:
\begin{itemize}
    \item The cosine similarities between Prompt A and the Prompt Type B prompts should be unimodal: there should be a single target length $l_B$ such that the activations from the Prompt Type B prompt with target length $l_B$ has maximum cosine similarity with the Prompt A activations.
    \item The cosine similarities should be linear: the activations from the prompt with Prompt Type A and length target $l_A$ should have the greatest cosine similarity with the activations from the prompt with Prompt Type B and target length $m l_A$, for some fixed multiplicative constant $m$. (This constant corresponds to the \enquote{conversion rate} between the unit that measures the Prompt Type A lengths and the Prompt Type B lengths.)
\end{itemize}

\begin{figure}
    \centering
    \includegraphics[width=\linewidth]{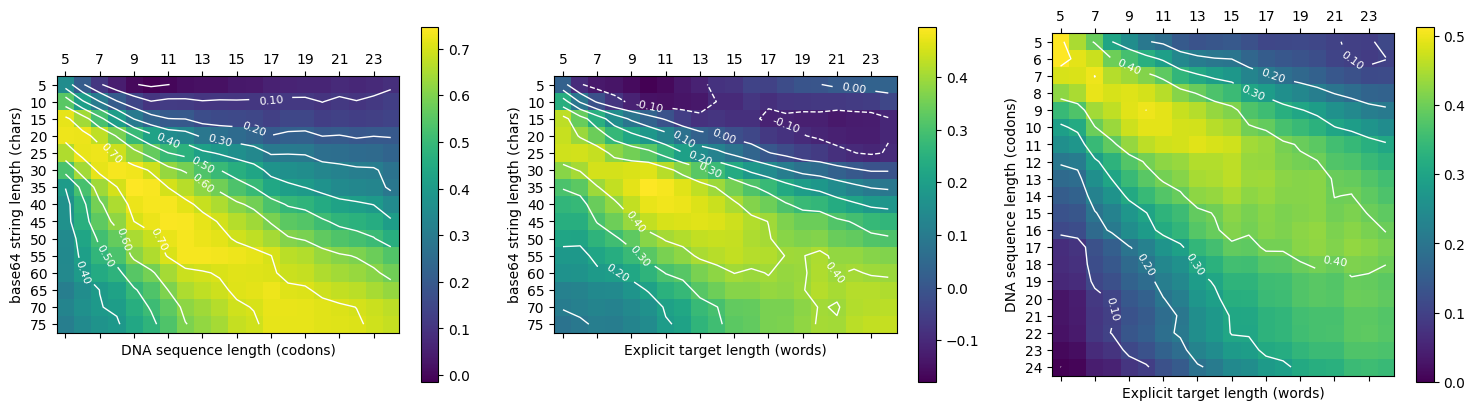}
    \caption{Cosine similarities between \enquote{length unit token} activations from prompts containing explicit length information and ending delimiter activations from prompts without explicit length information }
    \label{fig:probeCossims}
\end{figure}

\begin{figure}
    \centering
    \includegraphics[width=0.8\linewidth]{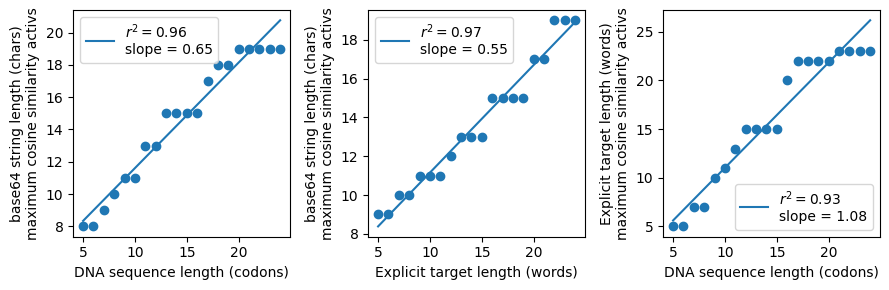}
    \caption{For different pairs of prompt types, the target lengths for one prompt type vs. the target length for the other prompt type whose activations have the greatest cosine similarity.}
    \label{fig:probeCorrelations}
\end{figure}

\paragraph{Results}
Cosine similarities are shown in Figure~\ref{fig:probeCossims}. We can see the unimodal pattern suggestive of a shared representation of target length information across different prompt types. Furthermore, in Figure~\ref{fig:probeCorrelations}, we see that there are in fact highly linear relationships ($0.93 \le r^2 \le 0.97$) between the target lengths for one prompt type and the target lengths of the activations from the other prompt type with the greatest cosine similarity. This strongly suggests that implicit length information is stored similarly to explicit length information, and that this information is stored similarly across tasks.

\subsection{Countdown geometry and universality}

We have seen that the countdown heads make use of length information be it explicitly or implicitly stored. But how do they do this? Looking at PCA plots and cosine similarities shows that the length information contained in countdown head key activations lies on a curved manifold, and that the current position information contained in countdown head query activations lies on a similar manifold. These manifolds are aligned such that the highest pairwise cosine similarities between the key activations and query activations are realized when the query activations come from the string position corresponding to the target length in the key activations. Notably, \textit{this is the same geometric structure identified in the line-wrapping subcircuit in Claude 3.5 Haiku}~\citep{gurnee2025when}. This implies that the countdown subcircuit discussed in this work is not merely used generally across tasks, but also exemplifies a motif that is general across models as well.

\subsubsection{Setup}

We use the randomly-generated base64 prompts described earlier in Section~\ref{sec:implicitCausal}. We generate prompts for strings with length between 5 and 49 characters inclusive.

All three of the countdown heads have the same KV head, so we obtain key activations from this single KV head at the ending delimiter of the second string in the prompt. For query activations, we store the activations for each of the three countdown heads at each position in the third string in the prompt.

\subsubsection{Cosine similarities: query positions are aligned with key lengths}

We compute cosine similarities for each countdown query head separately. Before computing cosine similarities, we center all key/query activations by subtracting the mean key/query activations respectively. Then, for each length and position, we compute the cosine similarity between the key activations corresponding to that length and each of the query activations corresponding to that position; we then take the mean over all positions for a given length.

\begin{figure}
    \centering
    \includegraphics[width=0.8\linewidth]{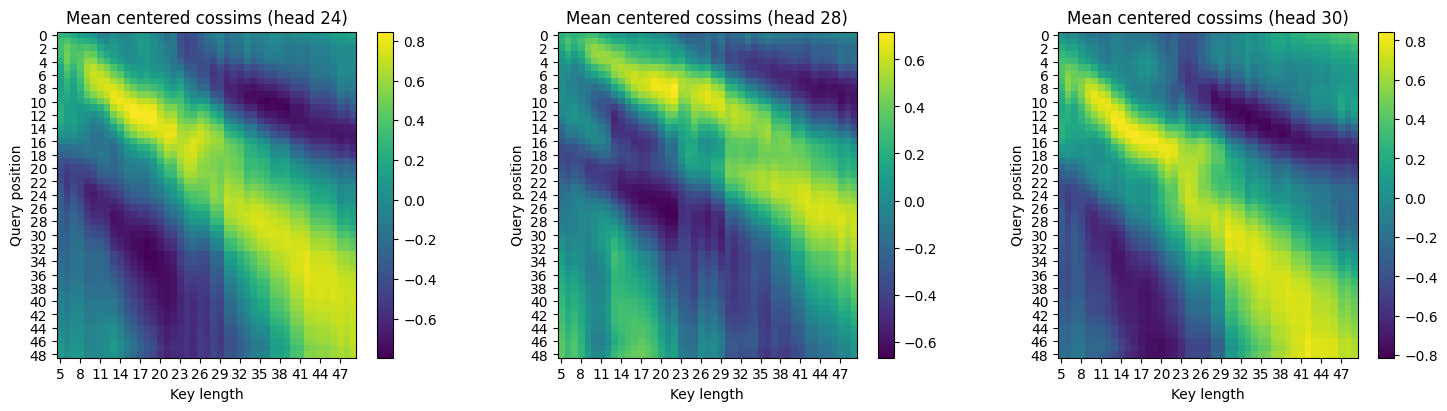}
    \caption{Cosine similarities between countdown head keys and queries.}
    \label{fig:countdownCossims}
\end{figure}

The cosine similarities are shown in Figure~\ref{fig:countdownCossims}. For heads 24 and 30, the highest cosine similarities are approximately on the diagonal, implying that the countdown heads detect that the target length is about to be reached when the current position is most similar to the target length. (For head 28, the pattern of highest cosine similarities is slightly skewed, with earlier queries attending to keys for longer lengths, corresponding to what we saw in Figure~\ref{fig:attnNorms}.) We also see a \enquote{ringing} or \enquote{banding} pattern where, moving from the diagonal to the corners, the cosine similarity drops, but then slightly rises again. This pattern was also found in the linebreak analysis of \citet{gurnee2025when}, who gave theoretical justifications for it (in particular, relating it to the properties of optimally projecting a higher-dimensional manifold into lower-dimensional space).

\subsubsection{PCA plots}

\begin{figure}
    \centering
    \includegraphics[width=0.45\linewidth]{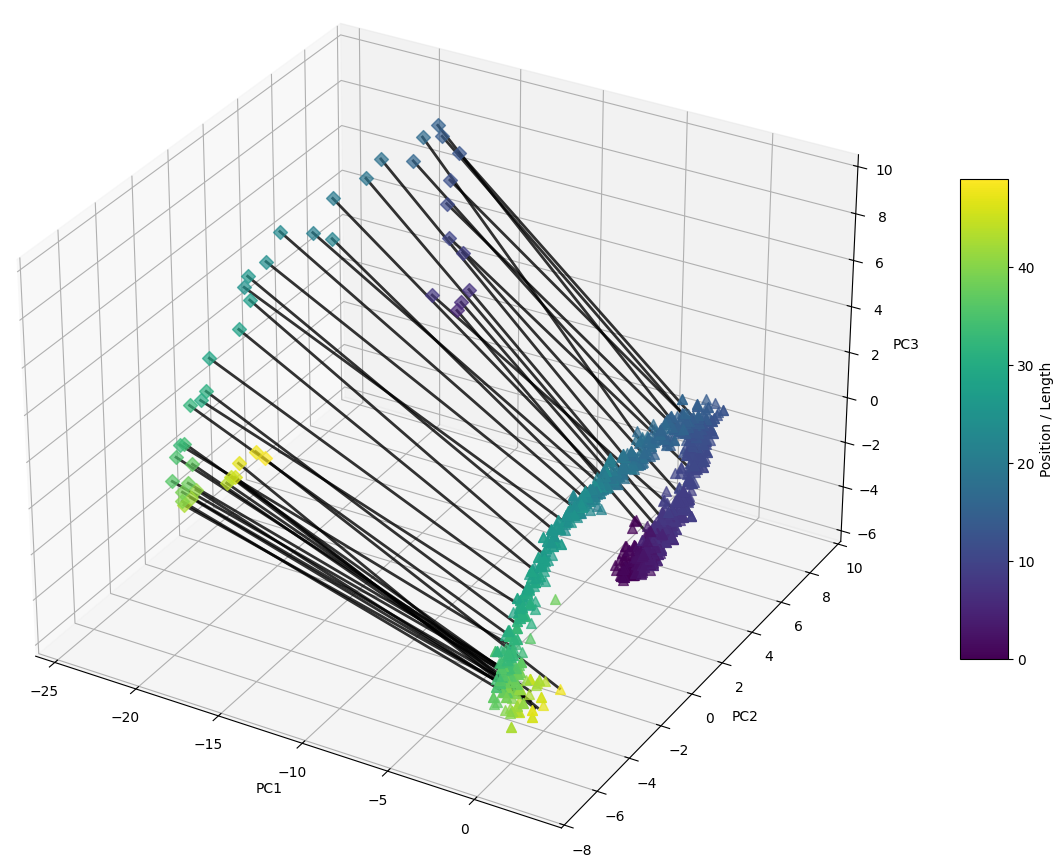}
    \caption{3D PCA visualization of countdown head key activations and query activations for different lengths. Color denotes target length (for keys) or string position (for queries); triangles denote keys and diamonds denote queries. Black lines connect each key activation to the mean activation of queries whose position in the string is the same as the key length.}
    \label{fig:pca}
\end{figure}

We also took the key activations and query activations and projected them onto their first three principal components. The resulting PCA plot can be found in Figure~\ref{fig:pca}. This plot further supports the finding from looking at cosine similarities that key activations for a given length are best-aligned with query activations whose position is the same as the key length. This geometric structure is the same described by \citet{gurnee2025when} used by models in line-wrapping fixed-width text, and further resembles general continuous 1D manifolds as described in \citet{wurgaft2026manifold} and \citet{engels2025not}.

\section{Task universality: using unsupervised probing to find more examples where the countdown subcircuit is used}
\label{sec:taskUniversality}
We have thus seen that the same countdown subcircuit is used across different tasks, including tasks where the target length is explicitly given in the prompt and tasks where it is implicitly suggested by the context.
Now, we will show that this subcircuit is used in a wide array of more general settings.

To do this, we constructed probes using the output of the countdown heads, ran these probes on a large corpus of text, and collected examples where the probe activated highly. The results of this unsupervised probing suggest that these attention heads are used by the model in a variety of contexts.

\subsection{Setup}

\paragraph{Probe construction} We construct probes using the outputs of the layer 18 \enquote{countdown heads} on the sentence planning prompts and completions. More specifically, we construct probes corresponding to a given \textit{distance from the end of the sentence}. We do this by running the model on many sentence planning prompts and their corresponding completions and storing the layer 18 attention output activations at all positions. Then, for the given distance from the end of the sentence, we compute the mean activations at that position over all sentences, and we subtract from this the mean activations at all other positions over all sentences. Thus, under the assumption that the output of the layer 18 \enquote{countdown heads} encodes distance from the end of the sentence (supported by Figure~\ref{fig:attnNorms}), each probe should fire highest at the corresponding distance from the end of the sentence.

\paragraph{Dataset and highly-activating example collection procedure}
We run our probes on a 10,000-sample subset of the Pile~\citep{gao2020pile}, where each sample is split into chunks of length 368. We compute probe activations by taking the dot product of each probe vector with the attention-out activations at each position in the chunk. For each chunk, we discard the probe activations at position 0, because the beginning-of-sentence token tends to have probe activations that are anomalously high. We then filter for chunks with at least one token whose probe activation is at least $1.0$ (a threshold chosen heuristically); these chunks constitute our \enquote{highly-activating examples}. We used a frontier LLM (Claude Opus 4.6) to summarize and label the highly-activating examples, and then manually selected ones that we found to be interesting. We also performed the same process on a 5,000-conversation subset of the LMSYS-1M chat dataset~\citep{zheng2024lmsys}, although we kept each conversation intact rather than splitting it into samples. 

\subsection{Survey of selected highly-activating examples}
Looking at examples where the probes had high activations, we found a diverse collection of themes, all of which involved the general behavior of the model \enquote{counting down} to the end of a sequence. The variety in these themes comes from the variety of cases in which this general countdown behavior is necessary. Thus, these examples illustrate the generality of the countdown heads identified earlier.

\begin{figure}
    \centering
    \includegraphics[width=\linewidth]{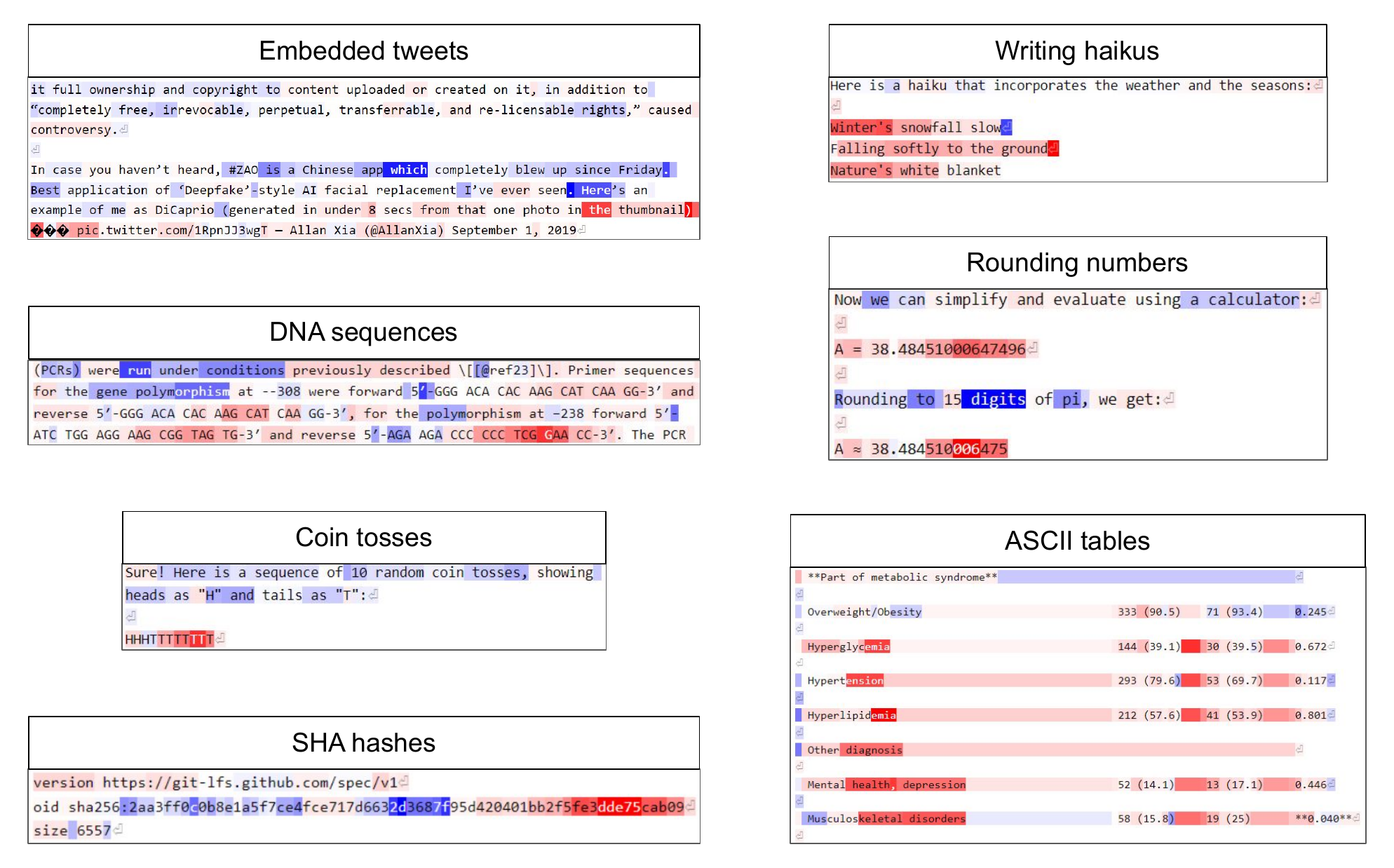}
    
    \caption{\textbf{(Zoom in to better see the examples.)} Examples \enquote{in the wild} where the countdown subcircuit is used, found by looking at highly-activating examples from the attention-out probes. Red denotes high positive activations, blue denotes very negative activations. The \enquote{coin tosses} and \enquote{rounding numbers} examples were found using a probe corresponding to \enquote{one word away from the end of the sentence}; all other examples were found using a probe corresponding to \enquote{two words away from the end of the sentence}. The \enquote{coin tosses}, \enquote{rounding numbers}, and \enquote{writing haikus} examples come from the LMSYS-1M chat dataset; all other examples come from the Pile.}
    \label{fig:probeExamples}
\end{figure}

Figure~\ref{fig:probeExamples} provides a selection of these high-activating examples. Many of these examples involve implicitly-determined length information, where the target length has to be inferred by the model, rather than being explicitly given in a prompt. Our selected examples are as follows:

\paragraph{Embedded tweets} For this theme, the probe fires on tokens at the very end of tweets embedded in news articles. Because most tweets have a maximum length of 280 characters, when predicting a tweet, there is thus a fixed length in characters to count down until. However, unlike the sentence planning example from earlier, the target length is not explicitly stated; instead, the model infers that it is reading a tweet, and \textbf{implicitly determines the target length}. What is particularly interesting is that in the case of tweets embedded in news articles, there is no explicit indicator of when the main body text ends and the tweet begins. Thus, the model has to first infer that it is reading a tweet before it can determine the target length.

\paragraph{DNA sequences of the same length} In this theme, the probe fires on tokens towards the end of a DNA sequence, in a prompt containing multiple previous DNA sequences of the same length. The model can infer the \enquote{target length} of the DNA sequence even though it is not explicitly stated, because all of the previous sequences were of the same length. Thus, this is another case where the model \textbf{implicitly determines the target length} from the prompt. Note that this example surfaced by probing was the inspiration for the setting investigated in Section~\ref{sec:mechanism}, demonstrating that unsupervised probing can effectively point towards fruitful directions of investigation.

\paragraph{Outputting a sequence of coin tosses} In this example, the model keeps track of how many characters are remaining until an explicitly-specified target length. Here, the characters (H and T) represent the outcomes of coin tosses. Note that although this example came from a chat dataset, the user's prompt did not specify any target length; instead, the target length was specified in the model's own response.

\paragraph{SHA-256 hashes} For this theme, the probe fires on tokens at the very end of an SHA-256 hash. Here, the length can be inferred from the prefix \code{sha256} at the beginning of the hash, which corresponds to 64 hexadecimal characters.

\paragraph{Writing haikus} This theme involves writing haikus, whose 5-7-5 syllable structure requires keeping track of both the current position within each line and the current line within the whole poem. The example shown reflects this, since not only does the probe fire on most tokens throughout the poem, but it fires the highest on the newline token after the second line, and has a \textit{negative} activation on the newline token after the first line (possibly as a result of the probe being extracted from activations two words away from the target length). This theme thus provides yet another example of the use of implicit length information.

\paragraph{Rounding numbers} This theme is similar to the \enquote{coin tosses} one. Here, the model rounds numbers to a given number of decimal digits. The example shown in the figure is a response to a request to compute the area of a circle and round to 15 digits of pi. Note that the model whose output is contained in the dataset interpreted this to mean rounding the result to approximately 15 digits, rather than first rounding pi to 15 digits and then computing the area. Nevertheless, probing Llama on this output shows that Llama is keeping track of how many three-digit tokens are remaining before the target length.

\paragraph{Aligned cells in ASCII tables} This theme of examples involves ASCII tables where cells are whitespace-aligned to have the same width. The probe fires on tokens towards the end of each cell. Similar to the DNA case, the information about the length of each cell is \textbf{implicitly determined} after the model sees the length of the cells in the previous row.

\section{Conclusion}

In this work, we reverse-engineered the single narrow task of outputting sentences of a given length, and found a countdown subcircuit that, as was revealed by unsupervised probing, is a general subcircuit, used in many different ways across many different tasks. Furthermore, we found that the model uses this subcircuit in conjunction with \enquote{implicit length information} inferred from context rather than explicitly stated.
Finally, we obtained evidence that this subcircuit is not merely universal across tasks within a single model, but also universal across models, since it makes use of the same geometric representations of length and position that Claude 3.5 Haiku does.

Thinking more broadly, we believe that our experience points towards an approach to carrying out mechanistic interpretability work wherein a practitioner begins with a specific task of interest, reverse-engineers a subcircuit responsible for that task, and then uses unsupervised methods to determine the broader generality of tasks that make use of that subcircuit. This approach can be viewed as an analogue, adapted to the circuits paradigm, of the common workflow when dealing with feature dictionaries such as SAEs or transcoders, wherein one first determines which features are active on a given input, and then considers the broader set of inputs on which these features activates. Notably, feature dictionary analysis has already seen great gains from automation \citep{bills2023language}, providing optimism that perhaps, this \enquote{bottom-up} approach to understanding subcircuits might also be amenable to automation. If so, then we are excited to see what other behaviors of LLMs can be reverse-engineered in this way.

\bibliographystyle{ACM-Reference-Format}
\bibliography{refs}

\end{document}